\def\year{2019}\relax
\documentclass[letterpaper]{article} 
\usepackage{aaai19}  
\usepackage{times}  
\usepackage{helvet}  
\usepackage{courier}  
\usepackage{url}  
\usepackage{graphicx}  
\usepackage{subfigure}
\usepackage{textcomp}
\usepackage[small]{caption}
\usepackage{amsmath,amssymb,amsfonts}
\usepackage{multirow}
\usepackage{booktabs}
\usepackage{paralist} 
\begin{document}
%
\title{Multi-Label Robust Factorization Autoencoder\\ and its Application in Predicting Drug-Drug Interactions  }
\author{Xu Chu, \ Yang Lin, \ Jingyue Gao, \ Jiangtao Wang, \ Yasha Wang, \ Leye Wang\\
\{chu\_xu, \ bdly, \  gaojingyue1997, \ jiangtaowang, \ wangyasha\}@pku.edu.cn, \ wly@cse.ust.hk}
\maketitle

\begin{abstract}
Drug-drug interactions (DDIs) are a major cause of preventable hospitalizations and deaths. Predicting the occurrence of DDIs helps drug safety professionals allocate investigative resources and take appropriate regulatory action promptly. Traditional DDI prediction methods predict DDIs based on the similarity between drugs. Recently, researchers revealed that the predictive performance can be improved by better modeling the interactions between drug pairs with bilinear forms. However, the shallow models leveraging bilinear forms suffer from limitations on capturing complicated nonlinear interactions between drug pairs. To this end, we propose \textbf{Mu}lti-\textbf{L}abel Robust \textbf{F}actorization \textbf{A}utoencoder (abbreviated to MuLFA) for DDI prediction, which learns a representation of interactions between drug pairs and has the capability of characterizing complicated nonlinear interactions more precisely. Moreover, a novel loss called CuXCov is designed to effectively learn the parameters of MuLFA. Furthermore, the decoder is able to generate high-risk chemical structures of drug pairs for specific DDIs, assisting pharmacists to better understand the relationship between drug chemistry and DDI. Experimental results on real-world datasets demonstrate that MuLFA consistently outperforms state-of-the-art methods; particularly, it increases $21.3\%$ predictive performance compared to the best baseline for top $50$ frequent DDIs. We also illustrate various case studies to demonstrate the efficacy of the chemical structures generated by MuLFA in DDI diagnosis.
\end{abstract}
\section{Introduction}

Drug-drug interactions (DDIs) are common situations in which a drug affects the efficacy and safety of another drug when both are administered together, resulting in many adverse drug reactions (ADRs) that may cause severe injuries or even be responsible for deaths \cite{Qato2016Changes}. Most DDIs are discovered by accident once a drug is already on the market \cite{Percha2013Informatics}. However, early detection of DDIs at preclinical stage based on data such as drug chemical structures helps drug safety professionals allocate investigative resources and take appropriate regulatory action \cite{Zhang2015Label}. In fact, predicting DDIs based on chemical structures is possible since the concept that similar chemical structures bring about similar biological properties has been employed over the years by medicinal chemists \cite{Traphagen2002Do,gedeck2008exploiting}. With the accumulation of massive adverse events data caused by DDI collected by systems such as FDA Adverse Event Reporting System (FAERS)\footnote{https://open.fda.gov/data/faers/}, using computational methods to predict DDI becomes feasible, and DDI prediction is drawing increasing attention of the AI research community.


Since ADRs (e.g., Nausea, Emesis, High blood pressure, etc.) associated with DDIs are important for both clinical and pharmaceutical decisions \cite{Vilar2017Detection}, we can classify DDIs into different types according to different ADRs in DDI prediction. 
The DDI prediction problem studied in this paper is defined as: Predicting the occurrence of different types of DDIs between a pair of drugs based on the drug features (e.g., chemical structures). 

In literature, similarity-based methods have been widely applied for DDI prediction \cite{Vilar2012Drug,Zhang2015Label,Abdelaziz:2017:LST:3165322.3165584,Kastrin2018Predicting}. These methods first calculate the similarities between each pair of drugs based on independently extracted drug features, and then based on those similarities, they predict the type of DDIs between drug pairs. The idea of the prediction is that, if drug A is similar to drug B, then the drugs that have DDIs with drug A are likely to have the same type of DDIs with drug B. Recently, researchers prove that the predictive performance can be improved by better modeling the interactions between drug pairs by bilinear forms \cite{jin2017multitask}. Despite the impressive results achieved with this approach, the question remains as to whether there is a better approach that could be used to capture the complicated nonlinear interactions between drug pairs more precisely.

Nowadays, deep representation learning methods has been found advantageous in modeling the complicated nonlinear relations \cite{krizhevsky2012imagenet,collobert2011natural}. DDI prediction is actually a multi-label classification problem. A natural idea is to harness the power of representation learning to learn a representation of interactions between drug pairs that is efficient for classification. With adequate labeled data, supervised methods are encouraged to represent the classes of DDIs in a linearly separable way. However, of all possible combinations of two drugs, only a small proportion of drug pairs are labeled with DDIs. With insufficient labeled data, the supervised algorithms suffer severe overfitting and would achieve low predictive performance. On the other hand, unsupervised algorithms such as autoencoders allow us to exploit information hidden in unlabeled data and therefore improve performance of DDI prediction. However, the representations learnt by unsupervised methods would in general entangle factors related to the types of DDIs with other class-irrelevant factors and therefore introduce undesired bias for DDI prediction.
The aforementioned analysis inspires us that if we manage to disentangle the categorization factors across all factors, then we may use a supervised learning signal to train the representation of categorization factors. At the same time an unsupervised learning signal could be employed to exploit hidden information from large unlabeled data, regularizing the supervised learning process and thereby enhancing the generalization of model by restraining overfitting. The single-label learning method factorization autoencoder (FAE) \cite{cheung2015discovering} introduced a dimerous representation. FAE considers the class label to be part of the representation and the remaining part encode the class-irrelevant factors. To disentangle the categorization factors across all latent factors, FAE introduced a mini-batch based cross-covariance loss termed XCov that penalizes the covariance matrix of each dimension in the class-relevant coding part and each dimension in the class-irrelevant coding part.

However, simply extending the vanilla FAE to high-dimensional multi-label situations such as DDI prediction would fail to achieve the best result. The reason is that XCov estimates the cross-covariance in every mini-batch separately. When the batch size is small, the cross-covariance estimator employed by XCov would result in gradient descent directions with large variance and thus hurt the performance. On the other hand, a large batch size method tends to converge to sharp minimizers of the training function and result in a degradation in the quality of the model as measured by the ability to generalize \cite{keskar2016large}. We introduce a novel mini-batch based robust cumulative cross-covariance loss CuXCov that approximates the full-batch statistics, which guarantees a more accurate estimation and allows for better classification performance as well as robust representations of interactions between drug pairs. 

The decoder of the autoencoder can be utilized as a feature generator. With designed fabricared inputs, the generator could output feature vector associated with specific category. In the context of DDI prediction, the generator could output vectors describing high-risk chemical structures associated with specific types of DDIs and therefore providing hints for drug research and development process.

In summary, our contributions are summarized as follows:
 

\begin{itemize}
	\item We proposed \textbf{Mu}lti-\textbf{L}abel Robust \textbf{F}actorization \textbf{A}utoencoder (called MuLFA) for DDI prediction. MuLFA inherits the puissant expressing power of deep neural network to characterize the complicated nonlinear interactions between drug pairs and is capable of leveraging hidden information in unlabeled data.
	\item We proposed a robust cumulative cross-covariance loss CuXCov that approximate the full-batch statistics, which is designed to effectively learn the parameters of MuLFA by disentangling categorical factors across latent factors and thus improving the classification performance.
	\item We construct a dimerous representation, with which we could generate high-risk chemical structures for specific types of DDIs, assisting pharmacists to better understand the relationship between drug chemistry and DDI and providing hints in drug research and development process.
	\item Experimental results on real-world datasets demonstrate that MuLFA consistently outperforms state-of-the-art methods; particularly, it increases $21.3\%$ predictive performance compared to the best baseline for top $50$ frequent DDIs. We also illustrate various case studies to demonstrate the efficacy of the chemical structures generated by MuLFA in DDI diagnosis.
\end{itemize}

\section{Related Work}
In literature, there has been a long line of studies in DDI prediction based on preclinical data. From the methodological perspective, the most representative DDI prediction methods are two-stage similarity-based methods. 
Firstly, drug features are extracted for each drug independently, based on which similarites are calculated for all drug pairs. Secondly, based on the idea that similar drugs are also biologically similar, different strategies are employed to predict DDIs based on the similarites between drugs, e.g., nearest neighbor method \cite{Vilar2012Drug}, label propagation method \cite{Zhang2015Label}, link prediction method \cite{Abdelaziz:2017:LST:3165322.3165584,Kastrin2018Predicting}. Recently, researchers  proved that the predictive performance can be improved by better modeling the interactions between drug pairs by bilinear forms \cite{jin2017multitask}. However, the shallow model being used suffers from limitations on capturing complicated nonlinear interactions between drugs pairs. 

Inspired by the success of deep representation learning methods \cite{krizhevsky2012imagenet,collobert2011natural}, we introduce a method to better capture the complicated interaction relationship between drug pairs as well as leveraging hidden information in unlabeled data.

Some previous works try to employ additional preclinical data, such as targets and enzymes, to enhance DDI prediction \cite{Cheng2014Machine,Takeda2017Predicting,Zhang2017Predicting}. However, such additional data are not always available for all drugs of interest \cite{Abdelaziz:2017:LST:3165322.3165584}, limiting the usage scope of those methods. In future, we will consider how to incorporate more preclinical drug features, if such extra data can be obtained.

\section{Preliminaries}
We first define some notations to prepare our method.	 
	\subsubsection{Definition 1. } \textit{Drug Chemical Structure Data} 
	
	The drug chemical structure data contains substructure profiles of $m$ drugs. Define set $\mathcal{A}$ as $\mathcal{A}=\{d_1, d_2, ..., d_m \}$. The feature vector describing chemical structure of drug $d_p$ is represented by a $l$-dimensional vector $\textbf{D}_p$.
	
	\subsubsection{Definition 2.} \textit{Chemical Structure Vector of a Pair of Drugs}
	
	Let $\mathcal{B}$ denote the set of all possible drug pairs in chemical structure data, i.e., $\mathcal{B}=\{(d_p,d_q)| d_p, d_q \in \mathcal{A}, 1\leq p<q\leq m\}$. The chemical structure vector $\textbf{d}_{pq}$ of drug pair $(d_p,d_q)$ is the concatenation of substructure profile vectors $\textbf{D}_p$ and $\textbf{D}_q$, namely, $\textbf{d}_{pq}^T$=$(\textbf{D}_p^T,\textbf{D}_q^T)$.
	\subsubsection{Definition 3. } \textit{DDI Data} 
	
	The DDI data contains $n$ drugs and $v$ types of DDIs. We denote $\mathcal{C}=\{\tilde{d}_1,\tilde{d}_2,...,\tilde{d}_n\}$ as the set of $n$ drugs and $\mathcal{R}=\{r_1,r_2,...,r_v\}$ as the set of $v$ types of DDIs. Each type of DDI event corresponds to a specific type of adverse drug reaction\footnote{ADRs caused by $2$ co-administered drugs rather than ADRs caused by a single drug.}. For a given type of DDI event $r_i \in \mathcal{R}, i=1,2,...,v$, the data only records credible drug pairs that could be the causing factor. We define set $\mathcal{D}$ as $\mathcal{D}=\{(\tilde{d}_p,\tilde{d}_q)| \tilde{d}_p, \tilde{d}_q \in \mathcal{C}, 1\leq p<q\leq n, \tilde{d}_p \ $and$ \ \tilde{d}_q $\ are reported to be associated with at least one\ $ r_i, i=1,2,\cdots,v\}$.  We denote the set of drug pairs associate with $r_i$ as $\mathcal{E}_i$, $\mathcal{E}_i \subseteq \mathcal{D},i=1,2,...,v$.

		\subsubsection{Definition 4. } \textit{Set of Labeled Drug Pairs and Set of Unlabeled Drug pairs}
			
	 We let set $\mathcal{D}$ be the set of labeled drug pairs for all $v$ tasks. And set $\mathcal{F}=\mathcal{B} - \mathcal{D}$ be the set of unlabeled drug pairs\footnote{In general, databases recording chemical structure collect as much information as possible. It is reasonable to assume $\mathcal{D} \subseteq \mathcal{B}$.}.
	 
		\subsubsection{Definition 5. } \textit{Set of Positive Samples and Set of Negative Samples for occurrence of the $i$-th Type of DDI $r_i$} 
%

Let set $\mathcal{E}_i$ be the set of positive samples and set $\mathcal{G}_i=\mathcal{D}-\mathcal{E}_i$ be the set of negative samples for task $r_i$.

	\subsection{Problem Statement} The problem of DDI prediction is formulated as follows. 
	
	\textbf{Input:}
	\begin{itemize}
		\item The set of chemical structure vectors $\{\textbf{d}_{k}\}_{k=1}^{|\mathcal{B}|}$,\footnote{ $|\mathcal{S}|$ denotes the cardinality of set $\mathcal{S}$. $\{e_i\}_{i=1}^L$ denotes a set and the index of element $e_i$ in set ranges from $1$ to $L$.} where $\textbf{d}_{k}=\textbf{d}_{pq}$ for some $p$ and $q$ such that$(d_p,d_q) \in \mathcal{B}$.
		\item DDI training data: $\{r_i\}_{i=1}^v$ and the corresponding sets of positive samples and negative samples, $\{\mathcal{E}_i, \mathcal{G}_i\}_{i=1}^v$.
	\end{itemize}
	\textbf{Output:} Predicted occurrence of $r_i$ of testing drug pairs for each type of DDI event $i=1,2,\cdots,v$.

%

\subsection{Factorization Autoencoder and XCov Loss}
Researchers have proposed a semi-supervised factorization autoencoder that could be used in single-label learning\cite{cheung2015discovering}. Specifically, given an input $\textbf{x}$ and its corresponding one-hot class label vector $\textbf{y}$ for a dataset $\mathcal{D}$, FAE learns the \textit{high-level representation} (the last-layer of the encoder)  in the form of concatenation of two vectors, i.e., $f_{\Theta}(\textbf{x})^T=(\hat{\textbf{y}}^T, \textbf{z}^T)$. FAE considers the class label to be part of the high-level representation of its corresponding input. Using class labels, FAE incorporates supervised learning to a subset of high-level representation, transforming them into observed variable $\hat{\textbf{y}}$. The remaining subset $\textbf{z}$ accounts for the remaining variation of dataset. To disentangle the categorization factors from other latent variables, FAE adds a mini-batch based cross-covariance loss (termed XCov). XCov loss prevents vector $\textbf{z}$ from encoding input variations due to class label by penalizing the covariance matrix of each dimension in the class-relevant coding part and each dimension in the class-irrelevant coding part.

\begin{equation}
L_{XCov}=\frac{1}{2}\sum\limits_{ij}[\frac{1}{N}\sum\limits_s(\hat{y}_i^s-\bar{\hat{y}_i})(z_j^s-\bar{z_j})]^2.
\end{equation}
$N$ is mini-batch size, and $\bar{\hat{y}_i},\bar{z_i}$ denote means over examples. $s$ is an index over examples and $i,j$ index feature dimensions. In our problem, FAE cannot be directly used for two reasons: (1) FAE is a single-label learning method. (2) XCov estimates cross-covariance separately in each mini-batch and could result in descent directions with large variance. To this end, we extend the FAE to address the two issues in next sections.

\section{Methods}
\subsection{Framework Overview}
\begin{figure}[!htbp]
\centering
	\includegraphics[scale=0.22]{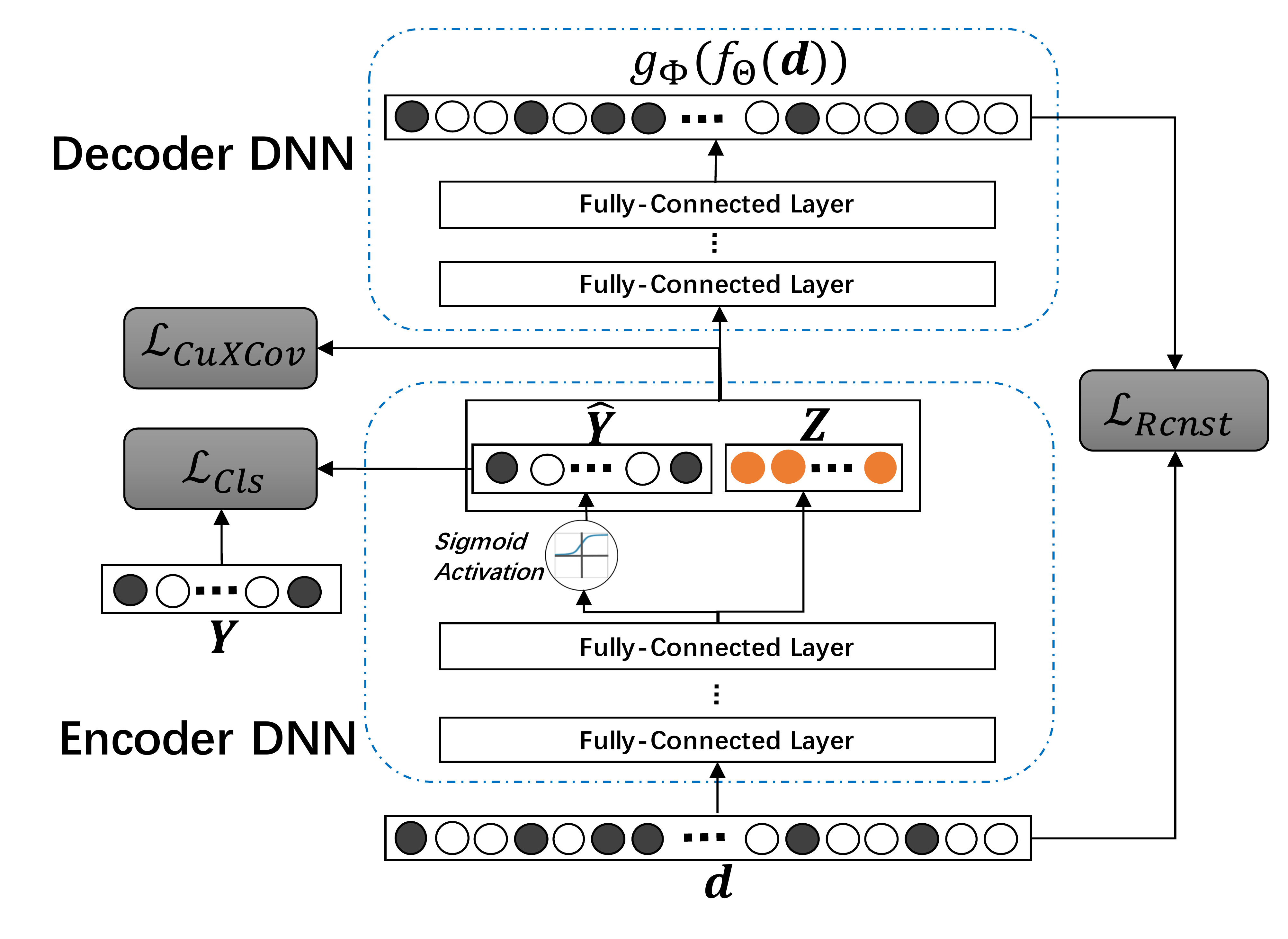}
	\caption{The architecture of MuLFA.}
	\label{fig:all}
\end{figure}

The overall neural network within MuLFA is built with an autoencoder structure. Figure \ref{fig:all} shows the architecture of our proposed method. 
The network consists of $H+1$ layers where $H$ is an even number. The first $\frac{H}{2}$ hidden layers are encoders to learn a representation of each input and the last $\frac{H}{2}$ hidden layers are decoders to reconstruct the input. 
For ease of illustration, we first fix some notations. Let $\textbf{L}^{(0)}=\textbf{d} \in \mathbb{R}^{2l} $ denote an input to the first layer and 
\begin{equation}
\textbf{L}^{(h)}= t^{(h)}((\textbf{W}^{(h)})^T\textbf{L}^{(h-1)}+\textbf{b}^{(h)}) \in \mathbb{R}^{d_h}
\end{equation}
be the output of the $h$-th layer, $h=1,2,\cdots,H$. $d_h$ denotes the dimension of the output at the h-th layer and $t^{(h)}$s are activation functions, which we take ReLU \footnote{We've tested other activations such as tanh and sigmoid by 10-fold cross-validation, among which ReLU performed best.} for all hidden layers except the high-level representation layer, i.e., for $h=\frac{H}{2}$,
\begin{equation}
\textbf{L}^{(\frac{H}{2})}=\begin{pmatrix}
\hat{\textbf{Y}}\\
\textbf{Z}
\end{pmatrix}=((\hat{y}_1,\cdots,\hat{y}_v)^T, (z_1,\cdots,z_u)^T)^T \in \mathbb{R}^{v+u}.
\end{equation} 
\begin{equation}
\begin{cases}

y_i=Sigmoid((\textbf{W}_i^{(\frac{H}{2})})^T \textbf{L}^{\frac{H}{2}-1}+b_i^{(\frac{H}{2})}), &i=1,2,\cdots,v; \\

z_i=(\textbf{W}_{v+j}^{(\frac{H}{2})})^T \textbf{L}^{\frac{H}{2}-1}+b_{v+j}^{(\frac{H}{2})}, &j=1,2,\cdots,u.

\end{cases}
\end{equation}
Where $\textbf{W}^{\frac{H}{2}}=(\textbf{W}_1^{(\frac{H}{2})},\textbf{W}_2^{(\frac{H}{2})},\cdots,\textbf{W}_{v+u}^{(\frac{H}{2})}) $ and $\textbf{b}^{\frac{H}{2}}=(b_1^{\frac{H}{2}},b_2^{\frac{H}{2}},\cdots,b_{v+u}^{\frac{H}{2}})^T$. The output of the top layer is $\textbf{L}^H$. We define the function of encoder as $f_{\Theta}$ and the function of decoder as $g_{\Phi}$. $\Theta$ and $\Phi$ denote the parameter space of encoder and decoder respectively.
\begin{equation}
f_{\Theta}(\textbf{d})=\begin{pmatrix}
\hat{\textbf{Y}}\\
\textbf{Z}
\end{pmatrix} \in \mathbb{R}^{v+u}, \ \ \ 
g_{\Phi}(f_{\Theta}(\textbf{d}))=\textbf{L}^H  \in \mathbb{R}^{2l}.
\end{equation}

Our goal of representation learning is to learn a representation $(\textbf{L}^{(\frac{H}{2})})^T=(\hat{\textbf{Y}}^T, \textbf{Z}^T)$ that isolates the categorization factors from other latent factors. $\hat{\textbf{Y}}$ codes the class labels of the input $\textbf{d}$ with $\hat{y}_i$ denoting the probability of the occurrence of the $i$-th event $r_i$. $\textbf{Z}$ codes the class-irrelevant factors to serve for semi-supervised learning by preserving as many factors of variation in the data as possible for the sake of reconstruction of input $\textbf{d}$.

For $i=1,2,\cdots,v$ , the learning of $\hat{y}_i$ can be viewed as a task $r_i$. In DDI prediction, each $\hat{y}_i$ is corresponding to a type of DDI event. Actually, different types of DDI events are related. For example, if a specific drug pair causes Nausea, then the specific drug pair is likely to cause Emesis. 
Thus the learning of $\hat{\textbf{Y}}$ can benefit from Multi-task learning for better exploiting the relatedness among tasks.  \cite{caruana1997multitask} characterized multi-task learning as an approach to inductive transfer that improves generalization and generalization error bounds \cite{baxter1995learning} by using the domain information contained in the training signals of related tasks as an inductive bias. Regarding outputs of layers $\{\textbf{L}_h\}_{h=1}^{H/2-1}$ as shared representation, hard under-sampled types of DDIs that could not be learnt in isolation are able to be learnt, and what is learnt for each task help other tasks be learnt better.
 
To factor the entangled source of variation relevant for categorization apart from other factors across the representation $\textbf{L}^{(\frac{H}{2})}$, We introduce a mini-batch based robust cumulative cross-covariance CuXCov loss to approximate the full-batch statistics, aiming at minimizing the entries in cross-covariance matrix of $\hat{\textbf{Y}}$ and $\textbf{Z}$ for all samples.

\subsection{Loss Function}
We now present details of how to train our model.
\subsubsection{Goal}
The training objective of  MuRFA is to minimize the weighted integration of 3 losses:
\begin{equation}
\min\limits_{\Theta, \Phi} \mathcal{L}_{Cls}+\beta \mathcal{L}_{CuXCov}+\gamma\mathcal{L}_{Rcnst}.
\end{equation}
Where hyperparameters $\beta>0$ and $\gamma>0$ control relative weights of $\mathcal{L}_{CuXCov}$ and $\mathcal{L}_{Rcnst}$ over $\mathcal{L}_{Cls}$. $\mathcal{L}_{Cls}$ penalizes the discrepancy between groundtruth labels and predicted occurrence probabilities of samples in labeled sets. $\mathcal{L}_{CuXCov}$ penalizes the estimated values of entries in cross-variance matrix of $\hat{\textbf{Y}}$ and $\textbf{Z}$. $\mathcal{L}_{Rcnst}$ is the general reconstruction error in autoencoders, penalizing the discrepancy between input $\textbf{d}$ and $g_{\Phi}(f_{\Theta}(\textbf{d}))$ for all training samples.

\subsubsection{CuXCov Loss}
When the number of different classes is large, (e.g., in DDI prediction, the known number of different classes of DDIs is more than 1,000.) the cross-covariance matrix between $\hat{\textbf{Y}}$ and $\textbf{Z}$ would in general require a large sample size to achieve an accurate estimation. However, the XCov loss in (1) estimates cross-covariance separately in each mini-batch and could result in descent directions with large variance. To address this issue, inspired by \cite{chang2018scalable}, we propose a cumulative loss CuXCov that approximates the full-batch cross-covariance matrix. This cumulative strategy  can trace back to \cite{Welford1962Note}, where the author proposed an accurate, one-pass, incremental approach to estimate the second central moment. Let $\Sigma^k_f,\Sigma^k_c,\Sigma^k_m,\Sigma^k_a$ denote the \textit{full, cumulative, mini-batch, approximate} cross-covariance estimator at the $k$-th training step respectively. The approximation works as follows:
\begin{equation}
\begin{cases}
\Sigma^k_c=\alpha\Sigma^{k-1}_c+\Sigma^k_m, &with \ \Sigma^{0}_c=\textbf{0}, \\
p^k=\alpha p^{k-1}+1,&with \ p^0=0.
\end{cases}
\end{equation}
Where $\alpha \in [0,1]$ is the decay rate. Then let 
\begin{equation}
\Sigma^k_a=\Sigma^k_c/p^k.
\end{equation}
$\Sigma^k_a$ would start converging to $\Sigma^k_f$ as $k$-th gets larger.

Let 
\begin{equation} 
\begin{bmatrix}
\tilde{\hat{\textbf{Y}}} \\
\tilde{\textbf{Z}}
\end{bmatrix}
\in \mathbb{R}^{(v+u) \times N}
\end{equation}
denote the high-level representation over a mini-batch with size $N$, where $\tilde{\hat{\textbf{Y}}}=(\hat{\textbf{Y}^1},\hat{\textbf{Y}^2},\cdots,\hat{\textbf{Y}^N}),$ and $\tilde{\textbf{Z}}=(\textbf{Z}^1,\textbf{Z}^2,\cdots,\textbf{Z}^N).$

We write mini-batch cross-covariance estimator in matrix form
\begin{equation}
\Sigma^k_m =1/N(\tilde{\hat{\textbf{Y}}}(\textbf{I}-\textbf{e}\textbf{e}^T))(\tilde{\textbf{Z}}(\textbf{I}-\textbf{e}\textbf{e}^T))^T\\
=1/N\tilde{\hat{\textbf{Y}}}\textbf{H}\tilde{\textbf{Z}}^T \in \mathbb{R}^{v \times u}.
\end{equation}
Where $\textbf{e} \in \mathbb{R}^N$ is a column vector with all entries being $1$,\ $\textbf{I}$ is the identity matrix, and $\textbf{H}=(\textbf{I}-\textbf{e}\textbf{e}^T)(\textbf{I}-\textbf{e}\textbf{e}^T)^T \in \mathbb{R}^{N \times N}$.
From (7), (8) and (10), we have
\begin{equation}
 \Sigma^k_a=1/p^k(\alpha\Sigma^{k-1}_c+1/N\tilde{\hat{\textbf{Y}}}\textbf{H}\tilde{\textbf{Z}}^T) \in \mathbb{R}^{v \times u} .
\end{equation}

Our goal is to minimize all entries in $ \Sigma^k_a$. We define  CuXCov loss as:
\begin{equation}
\mathcal{L}_{CuXCov}=trace( (\Sigma^k_a)^T \Sigma^k_a)/2.
\end{equation}
The gradient of CuXCov loss with respect to  $\tilde{\hat{\textbf{Y}}}$ and $\tilde{\textbf{Z}}$ is:
\begin{equation}
	\frac{\partial \mathcal{L}_{CuXCov}}{\partial \tilde{\textbf{Z}}}=\frac{\alpha}{N(p^k)^2}(\Sigma_a^{k-1})^T\tilde{\hat{\textbf{Y}}}\textbf{H}+\frac{1}{N^2(p^k)^2}\tilde{\textbf{Z}}\textbf{H}^T\tilde{\hat{\textbf{Y}}}^T\tilde{\hat{\textbf{Y}}}\textbf{H}.
\end{equation}
\begin{equation}
\frac{\partial \mathcal{L}_{CuXCov}}{\partial \tilde{\hat{\textbf{Y}}}}=\frac{\alpha}{N(p^k)^2}\Sigma_a^{k-1}\tilde{\textbf{Z}}\textbf{H}^T+\frac{1}{N^2(p^k)^2}\tilde{\hat{\textbf{Y}}}\textbf{H}\tilde{\textbf{Z}}^T\tilde{\textbf{Z}}\textbf{H}^T.
\end{equation}

\subsubsection{Classification Loss}
Our model can achieve multi-label learning. The learnt $\hat{\textbf{Y}}=(\hat{y}_1,\hat{y}_2,\cdots,\hat{y}_v)^T$ in representation for each input $\textbf{d}$ predicts the probabilities of occurrence of $r_i$ for $i=1,2,\cdots, v$. We use the labeled set for each task to supervise the training of $\hat{y}_i$s. 

\begin{equation}
\begin{aligned}
\mathcal{L}_{Cls}&=-\sum\limits_{\textbf{d}_s \in \mathcal{E}_i \cup \mathcal{G}_i}\sum\limits_{i=1}^v (\lambda y_i^slog(\hat{y}_i^s)+ (1-y_i^s)log(1-\hat{y}_i^s))
\end{aligned}
\end{equation} 
Where $i$ indexes tasks and $s$ indexes examples. $y_i^s$ is the label of $s$-th sample in labeled set with $1$ coding the occurrence of event $r_i$ and $0$ otherwise. $\lambda>1$ denotes the relative confidence of positive samples over negative samples. In the DDI prediction case, we record a drug pair associated with a type of DDI $r_i$ if the drug pair causes the corresponding adverse drug reaction. And no record of a drug pair associated with $r_i$ does not induce the conclusion that the drug pair would never cause the corresponding adverse drug reaction. Thus, it is more appropriate to treat positive samples and negatives samples discriminatively, and put larger weights on positive samples.


\textbf{Reconstruction Loss}
We often have a large amount of unlabled training data and relatively little labeled training data. To better explore information contained in unlabeled data, the architecture of autoencoder allows us to introduce the reconstruction loss over the whole training data.
\begin{equation}
\mathcal{L}_{Rcnst}=\sum\limits_{\textbf{d}_i \in \mathcal{B}} ||\textbf{d}_i-g_{\Phi}(f_{\Theta}(\textbf{d}_i))||^2.
\end{equation} 
Where $\mathcal{B}$ denotes the whole training set. Reconstruction loss also acts as a regularization term for the classification loss.

\subsection{Canonical Samples Generator}
After the auto-encoder is trained, 
\begin{equation}
\hat{\Theta},\hat{\Phi}= arg \min\limits_{\Theta, \Phi} \mathcal{L}_{Cls}+\beta \mathcal{L}_{CuXCov}+\gamma\mathcal{L}_{Rcnst}.
\end{equation} 
decoding function $G_{\hat{\Phi}}(\textbf{L}^H)$ can be used as sample generator given a high-level representation $\textbf{L}^{H_0}$. Formally, we define \\
\textbf{Definition 6.} \textit{The Canonical Sample with respect to the $i$-th Category:} Let $\hat{\textbf{Y}_i}$ be a vector with all entries being 0 except the $i$-th entry being 1. Let
\begin{equation}
\hat{\textbf{Z}}=E_{\mathcal{B}}(\textbf{Z}|\hat{\Theta},\hat{\Phi}),\ \ \ 
\textbf{C}_i=G_{\hat{\Phi}}
(\begin{pmatrix}
\hat{\textbf{Y}_i}\\
\hat{\textbf{Z}}
\end{pmatrix})
\end{equation}
We name $\textbf{C}_i$ the feature vector of the canonical sample with respect to the $i$-th category, for $i=1,2,\cdots,v$. The corresponding sample is named as the canonical sample with respect to the $i$-th category. \\

Take the last $u$ entries of $1/|\mathcal{B}|\sum\limits_{\textbf{d}_i \in \mathcal{B}}f_{\hat{\Theta}}(\textbf{d}_i)$
as $\tilde{\hat{\textbf{Z}}}$ to estimate $\hat{\textbf{Z}}$. Then an approximation of $\textbf{C}_i$ is $\tilde{\textbf{C}}_i=G_{\hat{\Phi}}(\begin{pmatrix}
\hat{\textbf{Y}_i}\\
\hat{\tilde{\textbf{Z}}}
\end{pmatrix}).$

In DDI prediction, when the inputs of MuLFA are chemical structure vectors of drug pairs, the canonical sample with respect to the i-th type of DDI event would be approximate canonical chemical structures of a pair of drugs that could jointly lead to the corresponding adverse drug reaction.

\section{Experiments}
\subsection{Datasets}
\subsubsection{DDI Data} The DDI data we use is from Twosides database\footnote{http://tatonettilab.org/resources/tatonetti-stm.html} \cite{tatonetti2012data}. It contains 645 drugs and 1318 types of DDIs, and in total 63473 drug pairs associated with DDI reports that makes the labeled set $\mathcal{D}$ with $|\mathcal{D}|=63473$.
\subsubsection{Drug Chemical Structure Data} The chemical structure features we use are extracted from Pubchem\footnote{https://pubchem.ncbi.nlm.nih.gov/} substructure fingerprint, and are binary coded as an 881-bit feature vector, each bit representing a Boolean determination of the presence of a substructure. For fair comparison, we only extract structure features of all drugs appeared in Twosides database, namely, $\mathcal{A}=\mathcal{C}$ and $|\mathcal{B}|=\binom {645} 2$.

\subsection{Experiment I \ \ \ \ \ \textit{On Classification Performance}} 		
		
\textbf{Methods for Comparison}

\textbf{Baselines} We compared our model with the following methods.
\begin{compactitem}
	\item Nearest Neighbor (NN) method in \cite{Vilar2012Drug}.
	\item Label Propagation (LP) method in \cite{Zhang2015Label}.
	\item Dyadic Prediction (DP) method in \cite{jin2017multitask}.
\end{compactitem}

\textbf{Variants of our methods} We also studied the effect of different components proposed in our method. 
The networks were trained by back-propagation via Adam optimizer.
\begin{compactitem}
 \item MuLFA: Our proposed model.
 \item MuLFA-$R$: Our proposed model without considering the reconstruction penalty. 
 \item MuLFA-$X$: Our proposed model without considering the cross-covariance penalty.
 \item MuLFA-$X^+$： Our proposed model without $\textbf{Z}$ in high-level representation, leading to no cross-covariance penalty.
\end{compactitem}


\subsubsection{Evaluations} We randomly selected 10\% of drugs and masked all DDIs associated with these drugs for testing in alignment with \cite{Zhang2015Label}. 
DDIs associated with drugs not in testing set are used for training all models and we use 10-fold cross-validation to tune all hyperparameters of different methods.
For testing data, we evaluate all methods on different collections of DDIs. For a given collection of DDIs, we randomly selected 50\% of the testing set for evaluation and repeated the selection-evaluation process for 50 times. We report the mean and standard deviation of the Area Under Precision-Recall Curve(AUPR) over 50 repetitions. DDI data is highly unbalanced with small positive sample sets and much larger negative sample sets for different types of DDIs. It was shown in \cite{davis2006relationship} that the area under Receiver Operating Characteristic curve (AUROC) is not appropriate for unbalanced data and metrics such as AUPR should be used instead. 

We adopt a different strategy in constructing the sets of negative samples compared with DR. DR took the complement of Twosides DDI interactions  $\{\mathcal{G}_i\cup\mathcal{F}\}_{i=1}^v$ as negative samples, while we took  $\{\mathcal{G}_i\}_{i=1}^v$ as negative samples and $\mathcal{F}$ as the unlabeled set in our method. We examined the drug pairs in the $\mathcal{F}$ and found some drug pairs should not be co-administered, e.g., Carbamazepine (ID=2554) and Isoniazid (ID=3767) are a pair of drugs in $\mathcal{F}$. Concurrent use of Carbamazepine and Isoniazid may result in increased carbamazepine exposure and increased risk of isoniazid-induced hepatotoxicity \cite{wright1982isoniazid}. For drug pairs in $\mathcal{F}$ constructed from Twosides, no credible data can be observed and we are not able to judge whether the other drug pairs in $\mathcal{F}$ interact or not, thus we take $\mathcal{F}$ as the unlabeled set. For all methods, we utilized DDI interactions from Twosides $\{\mathcal{E}_i\}_{i=1}^v$as positive samples.


\subsubsection{Results and Discussion}
\begin{table}[!htbp]
	\centering
	\caption{AUPR of MuLFA against Baselines.}\label{tab:performance}
	\resizebox{\columnwidth}{!}{
		\begin{tabular}{c|c|c|c}
			\toprule
			\textbf{Methods}& Top 50 DDIs & Top 51-100 DDIs & Top 101-150 DDIs\\
			\hline
    		NN& 0.367(0.0030) & 0.265(0.0023) & 0.224(0.0028)\\
			LP& 0.360(0.0031)&	0.254(0.0025)&	0.211(0.0029)\\
			DR& 0.375(0.0025)&  0.272(0.0019)&  0.233(0.0026)\\
			\hline
             MuLFA&  \textbf{0.455(0.0029)} &\textbf{0.313(0.0027)} &	\textbf{0.270(0.0033)}\\
			\bottomrule
		\end{tabular}
	}
\end{table}

\begin{table}[!htbp]
	\centering
	\caption{AUPR of MuLFA against its Variants.}\label{tab:performance}
	\resizebox{\columnwidth}{!}{
		\begin{tabular}{c|c|c|c}
			\toprule
			\textbf{Methods}& Top 50 DDIs & Top 51-100 DDIs & All DDIs\\
			\hline
			 MuLFA-$R$&0.424(0.0035)	& 0.293(0.0028)&	0.236(0.0023)\\
    		 MuLFA-$X$ &0.429(0.0028) &	0.294(0.0026)&	0.243(0.0025)\\
			 MuLFA-$X^+$ &0.423(0.0028) &	0.288(0.0029)	&0.241(0.0020)\\
             MuLFA& \textbf{0.455(0.0029)}& \textbf{0.313(0.0027)} & \textbf{0.278(0.0019)}\\
			\bottomrule
		\end{tabular}
	}
\end{table}

Table 1 and 2 compare the performance of the proposed method against competing methods and the variants of MuLFA on different of collections of DDIs. The Top $X_1-X_2$ DDIs in the table denotes the collection of $X_1$-th to $X_2$-th most frequent DDIs. The tables show that MuLFA consistently achieves higher AUPRs as compared to all competing methods at different settings. More concretely, from Table 1 we can see that DR and MuLFA outperform NN and LP consistently by better modeling the interactions between drug pairs. As the tasks become harder, namely, when the labeled sets become more and more unbalanced, the performances of similarities-based methods decay faster for failing to exploit the relatedness among tasks. Furthermore, our proposed method achieves improvement consistently over DR by up to $21.3\%$ because the deep neural network framework is able to capture the complicated nonlinear interaction relationship between drug pairs more precisely. Table 2 shows the performances of MuLFA and its variants. MuLFA-$R$ is a supervised model which fails to leverage the information contained in the large unlabeled set. MuLFA-$X$ outperforms MULFA-$X^+$ consistently demonstrating the effectiveness of including $\textbf{Z}$ for encoding the class-irrelevant factors. MuLFA outperforms MuLFA-$X$ significantly demonstrates the necessity of disentangle the categorization factors across all latent factors.

\subsubsection{Sensitivity Analysis}
\begin{figure}[!htbp]
\centering
	\includegraphics[width=6cm, height=3.4cm]{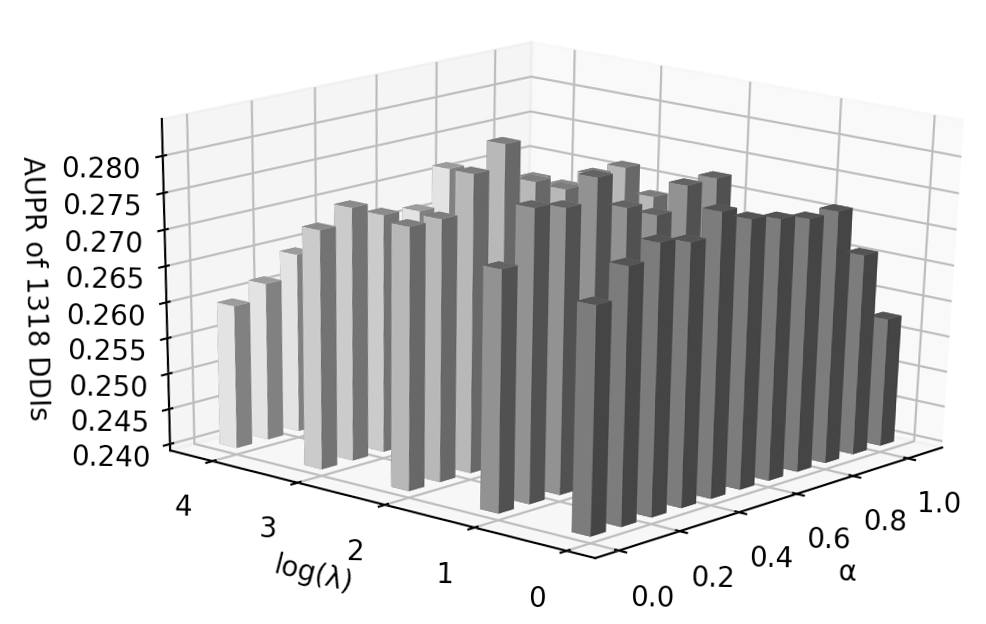}
	\caption{AUPRs of different combinations of $(\alpha,\lambda)$.}
	\label{fig:2}
\end{figure}
We study the sensitivity of two important parameters in our approach. The decay rate $\alpha$ in CuXCov loss and the relative confidence of positive samples over negative samples $\lambda$. We evaluate our model by grid search as shown in figure \ref{fig:2}, computing AUPRs of predicting 1318 DDIs simultaneously for different combinations of $(\alpha,\lambda)$. The best combination is $(\alpha,\lambda)=(0.3,4)$. Small values of $\alpha$ lead to estimations that lose too much information about early steps, while large values lead to too much emphasis on early steps and gain deficient information of a new batch. Small values of $\lambda$ make the model fail to discriminate the positive samples from less reliable negative samples, while large values of $\lambda$ prevent the model predicting an example being positive because of the high penalty. 

\subsubsection{CuXCov Loss vs XCov Loss}
\begin{figure}[!htbp]
\centering
	\includegraphics[width=8cm, height=3.4cm]{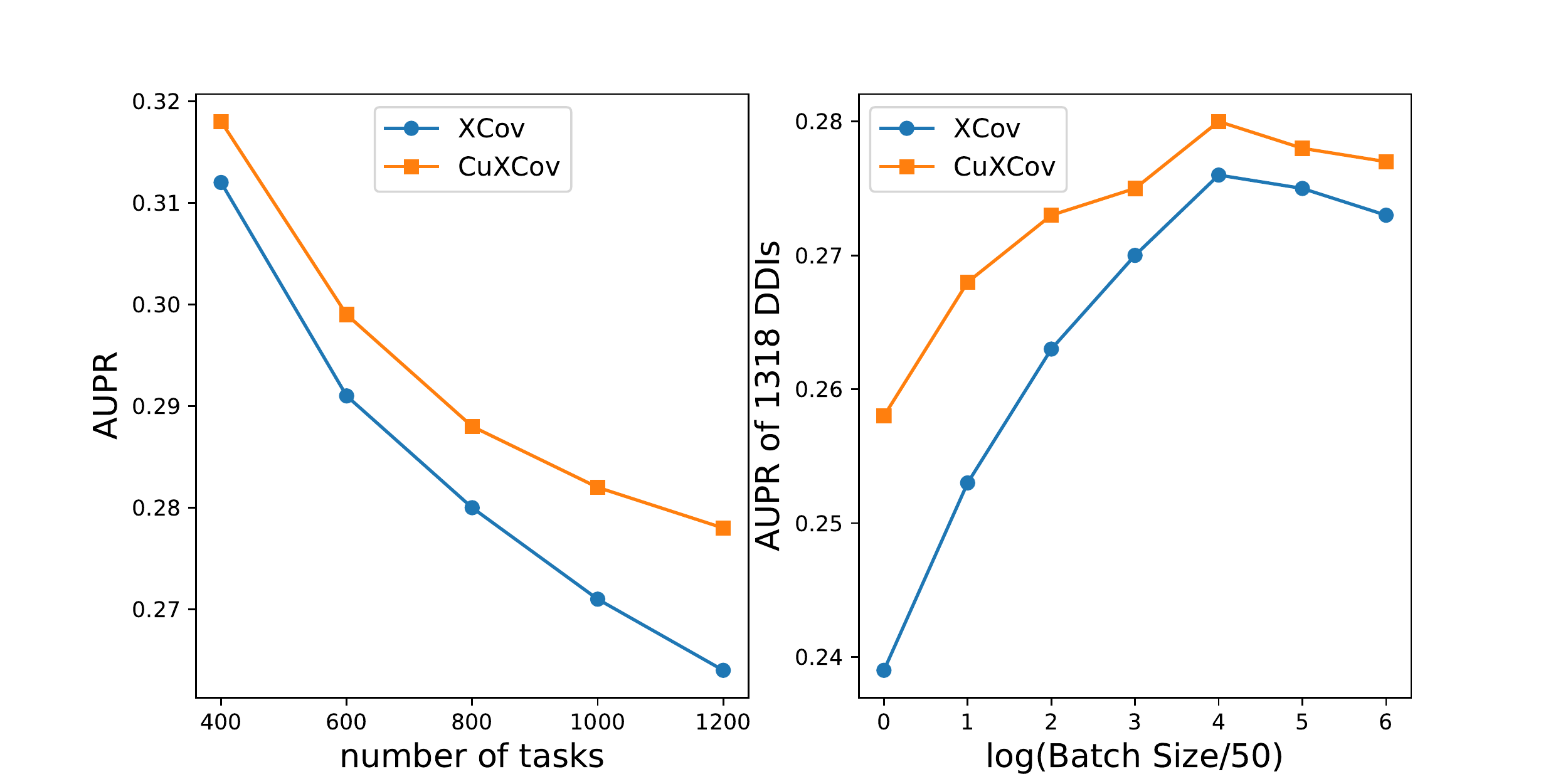}
	\caption{AUPRs of model leveraging CuXCov against model leveraging XCov loss in different settings.}
	\label{fig:3}
\end{figure}

We study the performances of models leveraging CuXCov loss against models leveraging XCov loss. Note that CuXCov loss degenerates to XCov loss when $\alpha=0$. For $\alpha=\{0,0.3\}$, we evaluate the predictive performances of 2 losses as shown in Figure \ref{fig:3}. In the left panel of Figure \ref{fig:3}, we study the influence of number of tasks on performance for mini-batch size being $200$. We evaluate the methods under different dimensions of $\hat{\textbf{Y}}$, i.e., the number of considered DDIs, from $400$ to $1200$ with step length being $200$. The result demonstrates that CuXCov model outperforms XCov model consistently and decays slower than XCov model as dimension of $\hat{\textbf{Y}}$ is getting higher. In the right panel of Figure \ref{fig:3}, we study the influence of batch size on performance for all $1318$ DDIs. We evaluate the methods under different batch sizes ranging from $50$ to $3200$. The result demonstrates that CuXCov model outperforms XCov model consistently. The performances of two models are increasing as batch size getting larger at first for better estimation of cross-covariance with smaller estimation variance. As the batch size continually getting larger, the performances of two models decay because of the degradation of generalization caused by convergence to sharp minimizers of the training function \cite{keskar2016large}.   

\subsection{Experiment II \\ \textit{Case Studies On Generated Chemical Structures}} 	
	\subsubsection{On Generated Canonical Chemical Structures of DDIs with High Co-occurrence}
Different types of DDIs are related. Usually, the ingenerate biological properties of DDIs with high co-occurrence are similar or even identical. For example, Sinus Tachycardia frequently co-occurs with Nausea and both of them can be activated by increased catecholamine release \cite{Koch1990Neuroendocrine}. Moreover, similar chemical structures may bring about similar medication effect mechanisms \cite{Traphagen2002Do}, which may cause DDIs. Thus it is reasonable to conjecture that the chemical structures of drug pairs that cause highly frequent co-occurred DDIs should be similar to each other. 
\begin{figure}[!htbp]
	\centering
		\includegraphics[width=8cm,height=3cm]{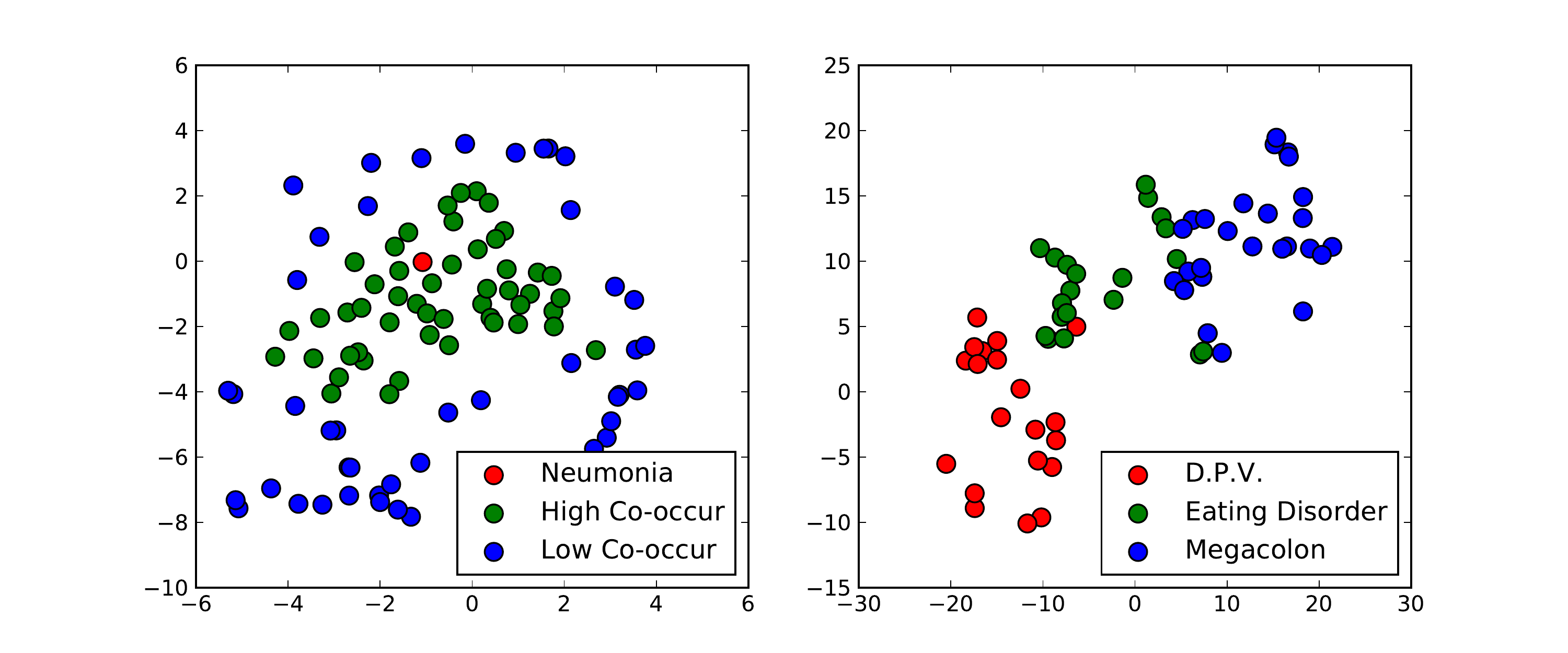}  
			\caption{Visualization of generated canonical chemical structure vectors of  some chosen DDIs by t-SNE.}\label{fig:4}    
\end{figure} 
The left panel of figure 4 is the $2$-dimensional t-SNE \cite{maaten2008visualizing} visualization of generated canonical chemical structure vectors of Neumonia, $50$ most frequent DDI types co-occur with Neumonia and $50$ least frequent DDI types co-occur with Neumonia. From the figure we can see that DDIs co-occur with Neumonia more frequently result in more similar generated canonical chemical structures. Megacolon, Eating Disorder and Disorder Perpheral Vascular are $3$ randomly selected types of DDIs with low co-occurrence with each other, and the right panel of figure 4 is the visualization of canonical chemical structure vectors of these 3 DDIs accompanied with 20 frequent co-occurred DDIs for each respectively. We can see 3 clusters in the embedding space in alignment with our conjecture.

	\subsubsection{On Explanation of Generated Canonical Structures}
For better interpretability we first constructed a reference set
 by randomly selecting 10\% of all drugs from Twosides database and masked all DDIs associated with these drugs. We used the  hyperparameters tuned in experiment I and trained MuLFA with the remain data. After the network is trained, we use the decoder to generate the canonical samples for top 20 DDIs with highest AUPRs. We consult pharmacy experts on the generated canonical structures. They confirmed that two classes of frequent generated high-risk structures 	are actually structurally similar to an inhibitor and a substrate of Cytochromes P450. Inhibitors and substrates are of high-risk to interact with other drugs. The pharmacy experts also confirm and explain the efficacy of the generated chemical structures of drug pairs with respect to DDI type Hypoventilation.

\textit{On Understanding the Frequent Generated High-risk Structures}
  We compare the feature vectors of the generated canonical samples with the vectors of drugs in the reference set, and we record the top 3 nearest neighbors for each generated drug structure. There are $120$ counts in total, among of which structures resembling Valproic Acid ($30/120$) and structures resembling Fentanyl ($25/120$) are  most frequent. 
	  	  
Cytochromes P450 (CYPs) are proteins of the superfamily that functions as an important enzyme system for drug metabolism. CYPs catalyze a wide range of oxidative reactions and are the most important pathway for drug metabolism. A drug can act as a substrate, an inducer or an inhibitor of CYPs. Inducers can increase the activity of the enzyme, and accelerate the metabolism of itself or other drugs. Inhibitors can attenuate the activity of the enzyme and slow down the metabolism of itself or other drugs. Inhibitors may also increase drug concentration that poses toxicity risks. Valproic acid is thought as an inhibitor of CYP450 2C9 (CYP2C9).  Co-administration of valproic acid with drugs that are primarily metabolized by CYP2C9 may result in increased drug concentration and adverse reactions would be observed. Fentanyl is the substrate of CYP3A4. Adverse events may occur if Fentanyl is co-administered with drugs that are the inducers or inhibitors of CYP3A4.

\textit{On Explanation of Generated Canonical Sample}
We compare the generated canonical feature vector of DDI type Hypoventilation with the drugs in reference set and find Ilopost and Venlafaxine are two most structurally similar drugs. From the chemical structure perspective, Iloprost($C_{22}H_{32}O_{4}$) is an eicosanoid, derived from the cyclooxygenase pathway of arachidonic acid metabolism, functioning as a stable analog of prostacyclin (PGI2). Iloprost is thought to promote benefit in pulmonary arterial hypertension (PAH) through vasodilation, antiproliferative effects, and inhibition of platelet aggregation\cite{Baker2005Inhaled}. Venlafaxine ($C_{17}H_{27}NO_{2}$) is a cyclohexanol and phenylethylamine derivative that functions as a serotonin-norepinephrine reuptake inhibitor (SNRI). \textit{In vitro} studies \cite{Sarma2010Venlafaxine} suggest that Venlafaxine would impact platelet aggregation. When Venlafaxine and Iloprost are administered together, Venlafaxine would decrease the drug efficacy of Iloprost by affecting the inhibition of platelet aggregation. Thus hypoventilation would be observed as a consequence of inadequate treatment of PAH.

\section{Conclusion}
In this paper, we propose a novel semi-supervised representation learning approach MuLFA for DDI prediction. We construct a dimerous representation for drug pairs, with which we can not only predict different types of DDIs simultaneously but also generate high-risk chemical structures for specific types of DDIs. We conduct extensive experiments on large-scale real-world data. The results demonstrate better classification performance of MuLFA than state-of-the-art prediction methods based on chemical structures. We also illustrate various case studies to demonstrate the efficacy of the chemical structures generated by MuLFA.


\bibliographystyle{aaai}
\bibliography{Paper_DDIPrediction}
\end{document}